\documentclass[conference]{IEEEtran}
\usepackage{CJKutf8}
\IEEEoverridecommandlockouts
\usepackage{cite}
\usepackage{amsmath,amssymb,amsfonts}
\usepackage{algorithmic}
\usepackage{graphicx}
\usepackage{textcomp}
\usepackage{xcolor}
\def\BibTeX{{\rm B\kern-.05em{\sc i\kern-.025em b}\kern-.08em
    T\kern-.1667em\lower.7ex\hbox{E}\kern-.125emX}}
\begin{document}

\title{Utilizing Social Media Attributes for Enhanced Keyword Detection: An IDF-LDA Model Applied to Sina Weibo\\
}

\author{\IEEEauthorblockN{Yifei Yue} \\
\IEEEauthorblockA{\textit{University of New South Wales} \\
\textit{Sydney, Australia} \\
\textit{z5392319@ad.unsw.edu.au}}
}

\maketitle

\begin{abstract}

With the rapid development of social media such as Twitter and Weibo, detecting keywords from a huge volume of text data streams in real-time has become a critical problem. The keyword detection problem aims at searching important information from massive text data to reflect the most important events or topics. However, social media data usually has unique features: the documents are usually short, the language is colloquial, and the data is likely to have significant temporal patterns. Therefore, it could be challenging to discover critical information from these text streams. In this paper, we propose a novel method to address the keyword detection problem in social media. Our model combines the Inverse Document Frequency (IDF) and Latent Dirichlet Allocation (LDA) models to better cope with the distinct attributes of social media data, such as the number of likes, comments, and retweets. By weighting the importance of each document based on these attributes, our method can effectively detect more representative keywords over time. Comprehensive experiments conducted under various conditions on Weibo data illustrate that our approach outperforms the baselines in various evaluation metrics, including precision and recall for multiple problem settings.

\end{abstract}


\section{Introduction}
With the rapid development of information technology, the amount of text information in social media has increased explosively. How to obtain critical information from massive text data has become more complex. A keyword is the smallest unit that can describe the topic of a document to a certain extent [1]. The discovery of representative keywords can essentially help one to understand the events hidden in massive data. Especially in this era of rapid development of social media such as Twitter and Weibo, keyword detection has been widely used to analyze the public interests and requirements, the latest trends [2], and even the time and place of natural disasters from recent text information released by the public [3]. In addition, the distinct attributes of social media, such as the number of likes, comments, and retweets, may also represent the importance of the document, and the detected keywords may also be more representative.

Existing methods on keyword detection can be divided into unsupervised keyword detection and supervised keyword detection. The unsupervised method does not require manually labeling the training set, which is faster, but it cannot effectively use a variety of features to rank candidate keywords. The supervised method can continuously adjust the influence of various information on the detection of keywords through learning, which has a better performance than the unsupervised method. However, the supervised method requires high labor costs, so the existing keyword detection mainly uses an unsupervised method with solid applicability. Many unsupervised keyword detection methods have been proposed. For example, in terms of frequency-inverse document frequency (TF-IDF), if a word appears in a document with high frequency and rarely appears in other documents, it is considered that the word has good classification ability. In addition [4], IDF is also a method that suppresses noise. Hofmann [5] proposed the PLSA model, which assumes every document contains a series of potential topics. Every word in the document is generated with a certain probability under the guidance of these topics. However, D.Blei et al. proposed that the aspect model used for PLSA has a serious overfitting problem. Therefore, they proposed the LDA model [6] based on PLSA, which can also be regarded as a combination of PLSA and Bayes' theorem. It applies Dirichlet priors to process document-topic and word-topic distributions, which helps to generalize better. Nowadays, the LDA model has also become one of the most commonly used topic models.

However, in terms of analyzing social media data, existing methods on keyword detection cannot use most of the features associated with the document, such as the number of likes, comments, and retweets. Therefore, in this paper, we look into the defects that models cannot utilize the unique features of social media. We propose the WBIDF-LDA model by combining IDF and LDA model to cope with Weibo attributes. The popularity of a weibo post is regarded as the sum of likes, comments, and retweets. We first use IDF to filter out the low frequency and exclusively high-frequency words in the original dataset, then select high-popularity weibo posts. Finally, we calculate the popularity of each weibo post according to relevant attributes and feed to the LDA model to extract the keywords under each topic.

We evaluate WBIDF-LDA with Weibo datasets. The results show that WBIDF-LDA outperforms the traditional LDA consistently and can detect more representative keywords. At present, the proposed WBIDF-LDA has a better effect on Chinese dataset.

\section{Our Method}

In this section, the proposed method, WBIDF-LDA will be explained.

\subsection{IDF filtering}

TF-IDF is a numerical statistic which is used to reflect the importance of a term to a document in a collection.

TF(Term Frequency), $tf(t, d)$, is the normalized frequency of term $t$ in document $d$:

\[tf(t,d)= \frac{f_{t,d}}{\sum\limits_{t' \in d} f_{t',d}}\]

where $f_{t,d}$ the number of times that term $t$ occurs in document $d$. 

From the equation, we know that a term could be important if it appears frequently in a document. Then the TF score will be high.

IDF(Inverse Document Frequency), is the frequency of its usage across documents:

\[idf(t,D)=  \log(\frac{N}{1 + |\left\{d \in D : t \in d|\right\}}) + 1\]

where N is the total number of documents in the corpus, i.e. $N = \left| D \right|$. 

From the equation, it shows that a term could be unimportant if it appears in many documents. Then the IDF score will be low since it is not a unique identifier.

When analyzing IDF separately, it can be found that IDF is able to filter out some words that appear in most documents. Moreover, IDF can also filter out uncommon words that only appear in a few documents, such as misspelled words, rare characters, or names. Therefore, IDF is applied to filter out these noise.

\subsection{Remove low popularity weibo posts}
The popularity of a weibo post is the sum of the number of likes, comments, and retweets as mentioned above. After filtering out much irrelevant noise, weibo posts with higher popularity are likely to be more important than lower ones because they are more likely to be approved by the public or more likely to trigger discussion. In contrast, weibo posts with low popularity tend to contain personal emotions and rarely describe important events. Therefore, to reduce noise, only high popularity weibo posts are selected for the following steps.

\subsection{Weight calculation}
Now a high popularity Weibo dataset is obtained after filtering out the noise mentioned above. However, the importance of each weibo post can be different due to their different popularity. Therefore, different weibo post needs to be weighted according to the corresponding popularity. Then, the weight is given by:

\[w =  1 + \log(1 + l) + 2\times\log(1+c)) + 4\times\log(1+r) \]

where w is the weight of a weibo post which is also the number of times that text content is duplicated, l, c and r are the number of likes, comments and retweets. Among these attributes, likes are often the easiest for weibo post to obtain, and retweets are often the most difficult for weibo post to obtain. Therefore, different attributes are also given different weights. The log function is used to avoid the weights of high popularity weibo posts being much larger than lower ones, which overwhelms the low popularity weibo posts.

\subsection{LDA modelling}

The LDA model is a three-layer Bayesian probability model, including a three-layer structure of words, topics, and documents. After the model parameters are obtained using Gibbs sampling, the document-topic and topic-word distributions can be produced by these parameters. This process can be demonstrated by Fig.1.

\begin{figure}[htbp]
   \centering
   \includegraphics[width=\linewidth]{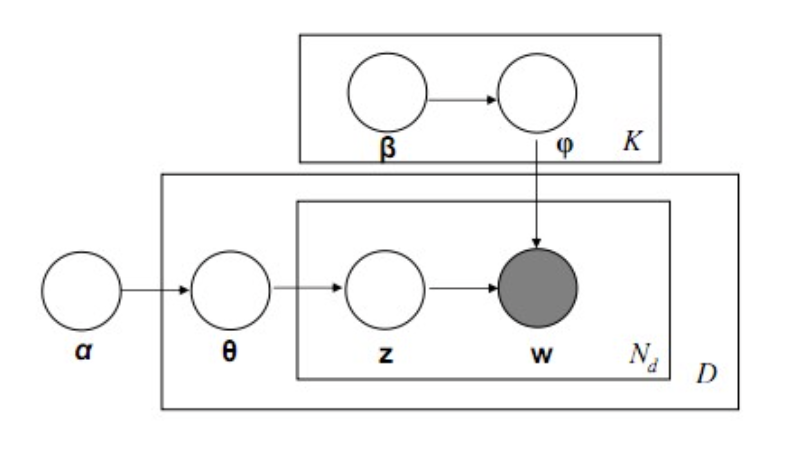}
    \caption{Bayesian network of LDA [8]}
    \label{fig:lda}
\end{figure}

The LDA model uses probabilistic generative models to model text datasets. For each document d in the dataset, LDA selects a multinomial distribution $\theta$ from the Dirichlet distribution $\alpha$, then a topic $z_{i,d}$ calculated by $\theta$ is assigned to a the $i$th word in the document. Second, a word $w_{i,d}$ is selected from the vocabulary based on the. $\phi$, which is the word distribution for each topic k. 

The joint probability distribution of all words in a document and the corresponding topic is given by:

\[p(w,z|\alpha,\beta)=\int_{\theta}\int_{\phi} p(w,z,\theta,\phi|\alpha,\beta)d\theta d\phi \]

The purpose of LDA modeling is to calculate the parameters $\theta$ and $\phi$, thereby obtaining the distribution of document-topic and topic-word. Gibbs sampling [10] can be used to construct a Markov chain that converges to the target probability distribution. Therefore, the estimated parameters of the sample can be obtained.

\section{Experiment}

In this section, WBIDF-LDA is evaluated on the Weibo dataset. Quantitative and qualitative analyses are conducted separately. The experiment results show that WBIDF-LDA outperforms the traditional LDA model consistently and can detect more representative keywords.

\subsection{Datasets}

18399 weibo posts related to Trump are collected from January 1st to January 31st, 2020. First, all characters except English, Chinese, and numbers are removed, and the dataset is also segmented by pkuseg toolkit [9]. The dataset at this time is named as Original dataset. Then, the top 4000 highest popularity weibo posts were selected which is named as 4000Hot dataset in order to prove that high popularity weibo posts have less noise and are more importance. 


In order to evaluate the effectiveness of detected keywords, the official Weibo trending is used as an important reference [7] for labeling the Original dataset. First, the top 20 hottest trends (hashtags) in January and 50 hashtags with the most occurrences in the dataset are selected. Then 70 weibo posts with high popularity and representative content under each hashtag are selected, and each weibo post is manually labeled by 3-5 keywords representing the content. Finally, the detection accuracy is calculated based on the overlap of detected words and labeled words.

\subsection{Evaluation Metrics}

To evaluate the accuracy of keyword detection on Twitter and Weibo datasets, precision, recall, and F1-score are applied.
\begin{itemize}
\item Successfully detected keywords: the intersection of labeled keywords and detected keywords.
\item Precision: the ratio of successfully detected keywords to all detected keywords
\item Recall: the ratio of successfully detected keywords to all labeled keywords, which is more important than precision because it represents the comprehensiveness of the results.
\item F1-score: 2 * precision * recall / (precision + recall)
\end{itemize}

Since recall represents the comprehensiveness of the results, it is more important than precision.

\subsection{Experiment Design}
\begin{CJK*}{UTF8}{gbsn}

By grid search, 30 is finally set as the number of topics for traditional LDA, IDF-LDA (No weight adjustment) and WBIDF-LDA. When the number of topics is too small such as 10, keywords of different events are more likely to appear under the same topic. When the number of topics is too large such as 70, the actual number of events is less than the number of topics, then more topics will contain useless keywords. It is difficult to detect representative keywords in both cases. The top 5 topics with the highest probability of different topic setting are shown in Table I and II below which demonstrates this situation. For example, when topic number is 10, the keywords in the first topic are “特朗普” (Trump), “华为” (Huawei), “世卫” (WTO), “疫苗” (vaccine) and “美国” (USA) which are not refer to the same event obviously. When topic number is 70, the keywords in the first topic are “美国” (USA), “计划” (plan), “和平” (peace), “承诺” (
promise) and “时间” (time). Based on these keywords, it is difficult to identify any specific event. Therefore, after multiple testing, when the number of topics is 30, LDA has the best performance of keyword detection. Then the above three methods are trained on the 4000Hot dataset. Finally, the top 10 words with the highest probability of each topic are selected as keywords for further comparisons.

\begin{table}[htbp]
\caption{Topic number $K = 10$}
\begin{center}
\begin{tabular}{|c|c|c|c|c|c|}
\hline
\textbf{Topic}&\multicolumn{5}{|c|}{\textbf{Keywords}} \\
\hline
\cline{2-5} 
\hline
1& 特朗普 & 华为 & 世卫 & 疫苗& 美国\\
\hline
2& 美国 & 特朗普 & 贸易 & 纽约& 关税\\
\hline
3& 白宫 & 印度 & 肺炎 & 确诊& 口罩\\
\hline
4& 经济 & 美元 & 接受 & 市场& 民调\\
\hline
5& 伊朗 & 美军 & 伊拉克 & 苏莱曼尼 & 1月\\
\hline
\multicolumn{6}{l}{}
\end{tabular}
\label{tab1}
\end{center}
\end{table}

\begin{table}[htbp]
\caption{Topic number $K = 70$}
\begin{center}
\begin{tabular}{|c|c|c|c|c|c|}
\hline
\textbf{Topic}&\multicolumn{5}{|c|}{\textbf{Keywords}} \\
\hline
\cline{2-5} 
\hline
1& 美国 & 计划 & 和平 & 承诺& 时间\\
\hline
2& 症状 & 特朗普 & 医院 & 治疗& 11日\\
\hline
3& 这就 & 代价 & 美国 & 生命& 西方\\
\hline
4& 美国 & 警察 & 特朗普 & 种族& 发生\\
\hline
5& 微信 & 禁令 & 行政 & 公司 & 45\\
\hline
\multicolumn{6}{l}{}
\end{tabular}
\label{tab1}
\end{center}
\end{table}

\end{CJK*}

\subsection{Experiment Result}

\begin{CJK*}{UTF8}{gbsn}

For qualitative evaluation, we compare the result of traditional LDA based on the Original dataset and 4000Hot dataset. The top 5 topics with the highest probability and also most likely to describe the same event are shown in Table III and Table IV.

\begin{table}[htbp]
\caption{Original dataset $K = 30$}
\begin{center}
\begin{tabular}{|c|c|c|c|c|c|}
\hline
\textbf{Topic}&\multicolumn{5}{|c|}{\textbf{Keywords}} \\
\hline
\cline{2-5} 
\hline
1& 特朗普 & 科比 & 新年 & 川普& 世界\\
\hline
2& 特朗普 & 中东 & 领袖 & 永远& 建议\\
\hline
3& 发表 & 特朗普 & 袭击 & 有人& 讲话\\
\hline
4& 弹劾 & 前往 & 特朗普 & 参议院& 弹劾案\\
\hline
5& 伊朗 & 特朗普 & 下令 & 美国 & 苏莱曼尼\\
\hline
\multicolumn{6}{l}{}
\end{tabular}
\label{tab1}
\end{center}
\end{table}

\begin{table}[htbp]
\caption{4000Hot dataset $K = 30$}
\begin{center}
\begin{tabular}{|c|c|c|c|c|c|}
\hline
\textbf{Topic}&\multicolumn{5}{|c|}{\textbf{Keywords}} \\
\hline
\cline{2-5} 
\hline
1& 科比 & 特朗普 & 去世 & 坠机 & 总统\\
\hline
2 & 中东 &特朗普& 北约 & 地区& 建议\\
\hline
3& 特朗普 & 发表 & 袭击 & 声明& 导弹\\
\hline
4& 弹劾 & 参议院 & 特朗普 & 众议院& 国会\\
\hline
5& 苏莱曼尼 & 暗杀 & 特朗普 & 美国 & 伊朗\\
\hline
\multicolumn{6}{l}{}
\end{tabular}
\label{tab1}
\end{center}
\end{table}

Although we can observe that the topics of the two datasets are describing similar events, the keywords detected in the original dataset have a lot of irrelevant information compared with the 4000Hot dataset, which are not representative. For example, topic 1 in original dataset describes the death of Kobe, but the keywords detected include “新年” (new year), “川普” (a nickname of Trump), “世界” (world) which are not relevant. In contrast, most of the keywords of topic 1 in Hot4000 dataset are more related to the event, which includes “科比”(Kobe), “去世” (death), “坠机” (air crash). In addition, the remaining topics in the two datasets have similar results. Therefore, we show that high popularity weibo posts have less noise and are more representative than lower ones.

For quantitative evaluation, we compare the results of traditional LDA, IDF-LDA and WBIDF-LDA running on 4000Hot dataset. From Table Ⅴ, we observe that WBIDF-LDA has the highest precision, recall and F1-score, and IDF-LDA is also outperforms LDA. These experimental results show that both the removing of low popularity weibo posts and IDF filtering can indeed filter out most of noise and improve the performance of keyword detection. In addition, the number of likes, comments and retweets also reflect the quality and importance of the corresponding weibo post.

\begin{table}[htbp]
\caption{A Larger precision recall or f1 score is better}
\begin{center}
\begin{tabular}{|c|c|c|c|}
\hline
\textbf{Model}&\textbf{Precision}&\textbf{Recall}&\textbf{F1-Score} \\
\hline
\cline{2-4} 
\textbf{LDA(Baseline)} & 0.322& 0.387& 0.352 \\
\hline
\textbf{IDF-LDA} & 0.346& 0.415& 0.377 \\
\hline
\textbf{WBIDF-LDA} & \textbf{0.389}& \textbf{0.464}& \textbf{0.432} \\
\hline
\multicolumn{4}{l}{}
\end{tabular}
\label{tab1}
\end{center}
\end{table}

\end{CJK*}

\section{Conclusion}

In this paper, we proposed WBIDF-LDA, a novel keyword detection method. In contrast to existing methods, WBIDF-LDA filters out a large amount of noise based on the popularity and IDF. Moreover, combining the unique attributes of Weibo, the weight of different Weibo will be adjusted according to the popularity. Finally, we demonstrated that WBIDF-LDA can effectively detect more representative keywords and topics in Weibo data after comprehensive experiments.
\\

\vspace{12pt}

\end{document}